\documentclass[journal]{IEEEtran}
\usepackage{amsmath,amsfonts}
\usepackage{algorithmic}
\usepackage{algorithm}
\usepackage{array}
\usepackage[caption=false,font=normalsize,labelfont=sf,textfont=sf]{subfig}
\usepackage{textcomp}
\usepackage{stfloats}
\usepackage{url}
\usepackage{verbatim}
\usepackage{graphicx}
\usepackage{cite}
\usepackage{float}
\usepackage{xcolor}
\usepackage{threeparttable} 

\usepackage{hyperref}
\hypersetup{
    colorlinks=true,
    linkcolor=blue,
    filecolor=magenta,      
    urlcolor=blue,
    }

\begin{document}

\title{A Fast Path-Planning Method for Continuous Harvesting of Table-Top Grown Strawberries}


\author{Zhonghua Miao†, Yang Chen†,~\IEEEmembership{Student Member,~IEEE,}, Lichao Yang, Shimin Hu, Ya Xiong*~\IEEEmembership{Member,~IEEE,}
\thanks{Manuscript received XXX, 2024; revised XXX, XXX. This work was supported by the Reform and Development Project of Beijing Academy of Agricultural and Forestry Sciences (BAAFS),  Haidian District Bureau of Agriculture and Rural Affairs, the BAAFS Innovation Ability Project (KJCX20240321, KJICX20240502), the BAAFS Talent Recruitment Program, and the NSFC Excellent Young Scientists Fund (overseas).(\textit{Corresponding Author: Ya Xiong})}
\thanks{Zhonghua Miao and Yang Chen contributed equally to this work and are co-first authors. They are with the School of Mechanical Electrical Engineering and Automation, Shanghai University, Shanghai 20044, China.}
\thanks{Yang Chen, Lichao Yang, Shimin Hu and Ya Xiong are with the Intelligent Equipment Research Center, Beijing Academy of Agriculture and Forestry Sciences, Beijing 100097, China (email: yaxiong@nercita.org.cn).}

\thanks{Personal use of this material is permitted. Permission from IEEE must be obtained for all other uses, in any current or future media, including reprinting/republishing this material for advertising or promotional purposes, creating new collective works, for resale or redistribution to servers or lists, or reuse of any copyrighted component of this work in other works.}
} %


\IEEEoverridecommandlockouts
\IEEEpubid{\makebox[\columnwidth]{978-1-6654-xxxx-x/25/\$31.00~\copyright~2025 IEEE. \hfill}%
\hspace{\columnsep}\makebox[\columnwidth]{ }}

\maketitle
\IEEEpubidadjcol

\begin{abstract}

Continuous harvesting and storage of multiple fruits in a single operation allow robots to significantly reduce the travel distance required for repetitive back-and-forth movements. Traditional collision-free path planning algorithms, such as Rapidly-Exploring Random Tree (RRT) and A-star (A*), often fail to meet the demands of efficient continuous fruit harvesting due to their low search efficiency and the generation of excessive redundant points.
This paper presents the Interactive Local Minima Search Algorithm (ILMSA), a fast path-planning method designed for the continuous harvesting of table-top grown strawberries. The algorithm featured an interactive node expansion strategy that iteratively extended and refined collision-free path segments based on local minima points. To enable the algorithm to function in 3D, the 3D environment was projected onto multiple 2D planes, generating optimal paths on each plane. The best path was then selected, followed by integrating and smoothing the 3D path segments. Simulations demonstrated that ILMSA outperformed existing methods, reducing path length by 21.5\% and planning time by 97.1\% compared to 3D-RRT, while achieving 11.6\% shorter paths and 25.4\% fewer nodes than the Lowest Point of the Strawberry (LPS) algorithm in 3D environments. In 2D, ILMSA achieved path lengths 16.2\% shorter than A*, 23.4\% shorter than RRT, and 20.9\% shorter than RRT-Connect, while being over 96\% faster and generating significantly fewer nodes. Additionally, ILMSA outperformed the Partially Guided Q-learning (QAPF) method, reducing path length by 36.7\%, shortening planning time by 97.8\%, and effectively avoiding entrapment in complex scenarios.
Field tests confirmed ILMSA’s suitability for complex agricultural tasks, having a combined planning and execution time and an average path length that were approximately 58\% and 69\%, respectively, of those achieved by the LPS algorithm.

\end{abstract}


\section{Introduction}
\IEEEPARstart{R}{esearch} and development in agricultural robotics has gained significant attention in response to increasing labor costs and shortages in traditional farming \cite{sinha2022recent}. Among the key technologies for robotic harvesting systems, path planning algorithms play a crucial role in enhancing the efficiency of fruit-picking robots \cite{karur2021survey}. These algorithms are essential for various functions, including obstacle avoidance, sequential planning, multi-robot coordination and obstacle separation\cite{xiong2020autonomous}. The performance of path planning not only determines the operational efficiency of harvesting but also affects the robot's adaptability to complex environments \cite{ren2020agricultural}.

\begin{figure}[!t]
\centering
\includegraphics[width=3.2in]{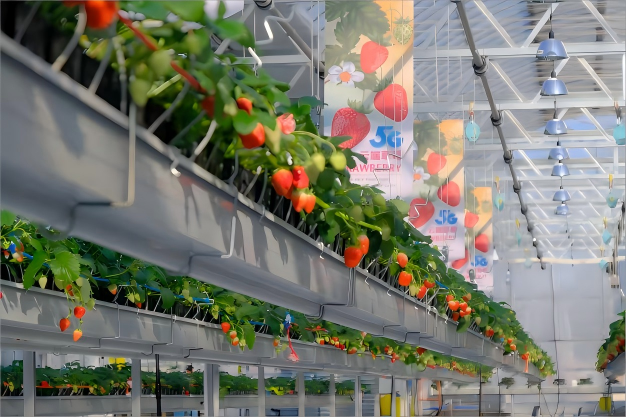}
\caption{Table-top grown strawberries picking scenario.}
\label{fig1}
\end{figure}

By continuously harvesting and storing multiple fruits in a single operation, robots can greatly minimize the travel distance needed for repetitive back-and-forth movements to pick and release the berries, thereby improving harvesting efficiency \cite{vrochidou2022overview}. Robotic grippers with fruit storage capabilities have enabled continuous harvesting, as demonstrated by the cable-driven gripper designed by Xiong et al. \cite{xiong2019development}, which includes a storage component. Additionally, swallowing-type harvesting grippers can achieve continuous harvesting by transferring the harvested fruits to a storage container \cite{kaleem2023development}. However, this also places higher demands on the real-time responsiveness, stability, and continuous planning capabilities of the path-planning algorithms. Furthermore, the complex growth environment of table-top grown strawberries, as depicted in Fig. \ref{fig1}, where fruit positions are random and intertwined with branches and leaves, poses additional challenges. The small size of strawberry stems makes them difficult for sensors to detect accurately, and the robotic arm must avoid colliding with the stems during operation. These constraints require that path-planning algorithms are robust to complex environments and generate collision-free paths \cite{ozturk2022review}.

Existing path-planning algorithms for obstacle avoidance include swarm optimization, artificial potential field methods, graph search algorithms, probabilistic roadmaps (PRM), and Rapidly-Exploring Random Trees (RRT) \cite{dai2022review}. While swarm optimization algorithms can find relatively optimal paths in complex spaces, they are prone to local optima and suffer from slow convergence \cite{polo2023multi}. Artificial potential field methods, such as those used in apple harvesting, although computationally efficient and suitable for real-time planning, are limited by potential field functions and struggle to find globally optimal paths in complex environments \cite{fan2020improved}. Graph search algorithms, such as A* algorithms, are powerful in finding optimal paths, but their efficiency decreases significantly in densely obstructed environments \cite{tang2021geometric}. PRM, like those used in citrus harvesting performs well in dynamic environments but tends to fail in narrow spaces due to sampling issues \cite{wu2021fast}. RRT is simple in structure and offers strong search capabilities, but the generated paths are often not smooth, and prolonged planning can result in unstable paths \cite{zeeshan2023performance}. Recent advancements in hybrid methods and machine learning-based planners have shown promise, but require and extensive parameter tuning and extensive training data\cite{xiao2022motion}. While these algorithms perform well in certain scenarios, they are not well-suited for the continuous harvesting of table-top grown strawberries in complex environments \cite{meng2022rrt}. Therefore, it is necessary to develop a fast path-planning algorithm for the continuous harvesting of table-top grown strawberries that takes into account the limitations of the visual system. This will enhance the harvesting efficiency and ensure damage-free picking  \cite{luo2024research}.

Additionally, it is necessary to design an efficient harvesting system capable of real-time monitoring of the robot's status and rapidly adjusting its motion path to accommodate the constantly changing environment \cite{peng2022strawberry}. Most existing systems focus on single-action or pick-and-place tasks, lacking the capability for continuous harvesting \cite{zhou2022intelligent}. For example, Parsa et al.  proposed an advanced modular autonomous strawberry-picking robotic system, but it still employs a single-fruit gripping and placing harvesting method \cite{parsa2024modular}.
In these system, advanced robotic arms with multi-degree-of-freedom capabilities have been developed to enable precise motion in constrained environments \cite{arikapudi2023robotic}. Vision systems such as RGB-D cameras and stereoscopic sensors allow for fruit localization and obstacle detection. However, they face challenges with occlusion caused by overlapping leaves and stems. Additionally, the integration of perception, planning, and harvest sequence often suffers from poor coordination, resulting in inefficiencies and suboptimal performance in dynamic environments \cite{li2023multi},\cite{harman2022multi}.

The main contribution of this paper is the proposal of a fast path-planning method applied to continuous harvesting of table-top grown strawberries. Its main novelties are outlined as follows:

1) An interactive node expansion strategy was proposed that iteratively extended collision-free path segments based on local minima point, balancing global and local optimization with low computational load and demonstrating strong real-time performance and adaptability.

2) Through the projection of the 3D environment onto multiple 2D planes, combined with collision detection and path smoothing techniques, the method refined potential paths to generate a smooth, collision-free trajectory, significantly enhancing path quality in complex, high-dimensional environments.

3) The successful deployment of this algorithm to strawberry-harvesting robots, combined with a control system, had verified its excellent continuous planning capabilities in harvesting tasks.

The rest of this paper is organized as follows. In Section II, we begin by introducing the path-planning problem in table-top grown strawberries, focusing on perception constraints and the limitations of existing planning algorithms. Then, we describe the harvesting system that supports the proposed path planning algorithm. In Section III, we present a detailed explanation of the proposed path-planning algorithm. In Section IV, we discuss the experimental results and performance evaluations conducted in both simulation and field environments. Finally, Section V concludes this paper with key findings and future research directions.

\section{Problem definition}
This section introduces the problem definition that motivated the development of our new path-planning algorithm.
Specifically, it focuses on the perception constraints imposed
by table-top grown strawberries and the limitations and challenges encountered when deploying existing algorithms in this context.
These challenges lay the foundation for proposing a more effective algorithm in subsequent sections.

\subsection{Perception Constraints}
\begin{figure}[t]
\centering
\includegraphics[width=3.3
in]{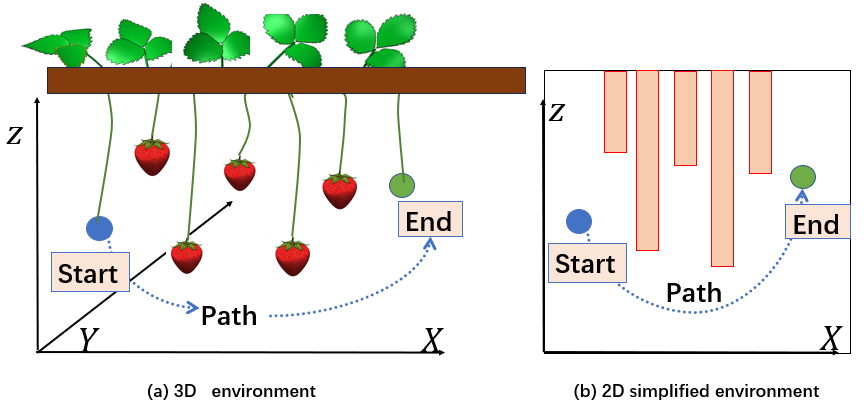}
\caption{Environment for planning the picking path of table-top grown strawberries: (a) 3D environment, (b) 2D simplified environment. }
\label{fig4}
\end{figure}

In the table-top grown strawberry-harvesting environment, the main challenge arised from the thin stems (typically 1-2 mm), which were difficult to detect accurately \cite{ge2023three}. As shown in Fig. \ref{fig4}(a), the strawberry positions were random and intertwined with stems and leaves. Although the fruit itself was typically detectable, the fragile stems must not be collided with the gripper during harvesting. Thus, the robotic arm must maintain a safe distance from both the fruit and the stems and cannot pass through these delicate stems, imposing significant constraints on continuous harvesting path planning. 

\subsection{Limitations of Conventional Obstacle Avoidance Algorithms}

Several conventional algorithms have been applied to table-top grown strawberries, but they exhibit certain limitations. These include experience-based methods and the 3D Rapidly-exploring Random Tree (3D-RRT) algorithm in three-dimensional environments, as well as the Rapidly-exploring Random Tree (RRT), RRT-Connect, and A* algorithms in two-dimensional environments.

In the 3D environment (Fig. \ref{fig4}a), the picking arm must avoid collisions with both the stems and other strawberries  \cite{zhang2023design}.
We first implemented an experience-based approach, using the lowest point of the strawberry (LPS) as the critical point. Such special points used for obstacle avoidance would be referred to as key nodes thereafter. Besides, general trajectory points generated along the planned path would be referred to as nodes.
The path first moved vertically to this node’s height to avoid obstacles, then moved horizontally beneath the strawberry, executing the picking action with an upward motion. This path-planning strategy offered good real-time performance, but required longer paths to avoid obstacles, thus increasing execution time. Besides, we deployed an improved 3D-RRT algorithm, which initially directed the search towards the target plane before sampling \cite{martinez2022fast}. While this accelerated target acquisition, it often resulted in collision-prone paths. When combined with bidirectional tree expansion, its real-time performance was improved, but the final paths were suboptimal, and the trees often got stuck in local minima \cite{xin2023improved}.

By reducing the complexity of 3D data to a 2D plane, we simplified the strawberry-picking environment, as shown in Fig. \ref{fig4}(b), allowing for more efficient path searching in this complex task. In the 2D environment, RRT algorithm quickly covered the picking space
\cite{ganesan2024hybrid}. However, its paths contained excessive redundant points, and as the environment became more complex, node expansion times increased, reducing search efficiency. Although we implemented an improved RRT-connect algorithm, the computational cost remained high and path smoothness issues persisted \cite{kang2021improved}. Additionally, the A* algorithm, leveraging an Euclidean heuristic, generated smoother, more feasible paths \cite{zhang2023global}. Yet, its performance was highly dependent on the heuristic’s accuracy, with diminished efficiency in complex or high-dimensional settings.

Overall, conventional algorithms failed to meet the demands of computational efficiency, path quality, and stability. While experience-based LPS offered better real-time performance, it resulted in redundant paths. A new algorithm was urgently needed to integrate the strengths of these methods and achieved fast path planning for continuous harvesting of table-Top grown strawberries.

\subsection{Challenges with Advanced Path-Planning Methods}

Although hybrid methods and machine learning-based planners have shown promise for complex robotic tasks \cite{liu2022new}. When applied to specific agricultural tasks, such as continuous harvesting of strawberries, these advanced Path-Planning Methods face significant challenges.

 Hybrid methods, such as those based on membrane pseudo-bacterial potential fields, combine membrane computing, the pseudo-bacterial genetic algorithm, and the artificial potential field method\cite{orozco2019hybrid}. These approaches can improve execution time and provide superior path planning solutions for autonomous mobile robots\cite{orozco2019mobile}. But they require extensive parameter tuning and incur high computational costs in dense or dynamic environments. These challenges limit their real-time applicability in agricultural tasks.
 
 More recent approaches have integrated machine learning techniques into path planning, particularly Q-learning, which enables self-learning without prior environmental models \cite{low2022modified}. However, Q-learning suffers from slow convergence to optimal solutions and can be inefficient in complex environments. To address these limitations, Partially Guided Q-learning (QAPF) combining  Q-learning with the artificial potential field (APF) method, has been applied to improve efficiency by guiding the agent toward optimal paths more quickly \cite{orozco2022mobile}. However, QAPF still faces challenges in complex, unstructured environments, where substantial training data are required, and the potential field may not always offer optimal guidance \cite{zhou2024optimized}.
 Recent modifications, such as integrating deep reinforcement learning, aim to further enhance adaptability and speed, but these approaches also introduce increased computational complexity and training time\cite{low2023modified}.

While existing hybrid and machine learning-based methods have made significant advances, their limitations in complex parameters, data requirements, and computational efficiency render them unsuitable for real-time continuous path planning in table-top strawberry harvesting. Additionally, while these advanced path-planning methods are primarily used in autonomous mobile robots, applying them to robotic arms for harvesting presents significant challenges and requires extensive adjustments.

\section{system design}
Prior to developing a new path-planning algorithm for continuous harvesting table-top grown strawberries, it was necessary to develop a continuous harvesting system. This section will provide a detailed overview of the system architecture and the visual perception.  

\subsection{Continuous Harvesting System}
\begin{figure}[!t]
\centering
\includegraphics[width=2.9in]{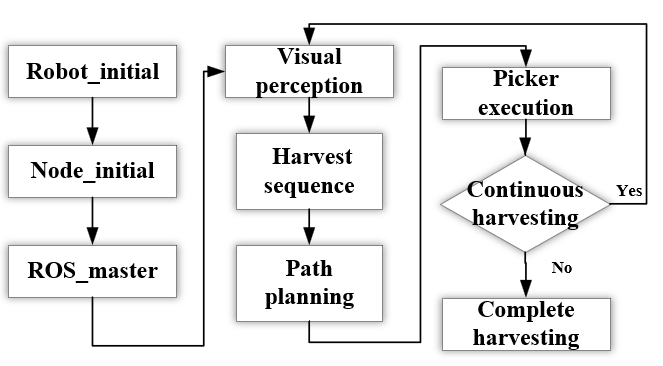}
\caption{Continuous harvesting control system.}
\label{fig2}
\end{figure}

In strawberry harvesting, sequentially harvesting multiple strawberries and temporarily storing them within the gripper can significantly improve picking efficiency. To achieve this, we developed a continuous harvesting system based on the robot operating system (ROS), which included visual perception, harvest sequence, and a path-planning algorithm, as shown in Fig. \ref{fig2}. Upon obtaining the positional information of the strawberries, the strawberry harvest sequence was allocated from bottom to top, which helps minimize the time spent on multiple task rerouting by the picking arm \cite{wang2024assisting}. After identifying the priority strawberries for harvesting, the arm followed an optimal, collision-free path guided by the path-planning algorithm to complete the harvesting process. Then the arm directly proceeded to the next picking cycle from the position of the previous fruit, re-engaging in perception, harvest sequence, and path-planning without returning to the fruit placement position, thereby achieving the task of continuously picking multiple strawberries. 

\subsection{Visual Perception}

\begin{figure}[t]
\centering
\includegraphics[width=3 in]{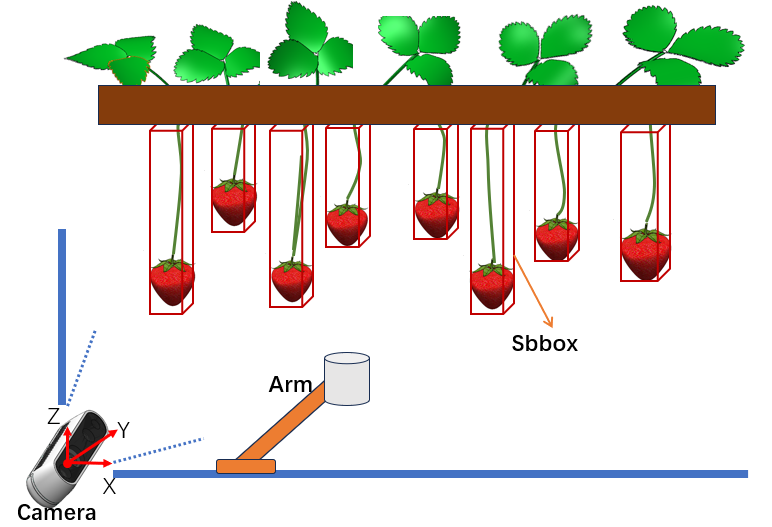}
\caption{Visual perception scenario.}
\label{fig3}
\end{figure}

Visual perception is a key module in the continuous harvesting system. The details are as follows. First, using a D455 depth camera, we captured both color and depth images \cite{beyaz2022accuracy}. Then, we employed the YOLOv8 algorithm for real-time detection and localization of strawberries \cite{khow2024improved}. The algorithm accurately identified the strawberry positions and outputted their three-dimensional coordinates, which were then converted into 3D bounding boxes (sbbox). To address the challenge of the robot end-effector avoiding thin strawberry stems, we extended the sbbox vertically to envelop the entire stem, as shown in Fig. \ref{fig3}. These sbboxes provided crucial input for path planning and obstacle avoidance. By processing the sbbox data, we accurately estimated obstacles boundaries, ensuring the robotic arm can avoid both stems and fruit, enabling safe and efficient picking.


\section{Interactive Local Minima Search Algorithm}

This section will introduce our newly proposed interactive local minima search algorithm (ILMSA) for continuous harvesting of table-top grown strawberries. The following will detail the implementation process of this algorithm, including the generation of new nodes, collision detection, obstacle projection, spatial path optimization, and the overall execution of the algorithm.

\subsection{ILMSA in 2D}
Our algorithm was initially developed in 2D and then this section first introduces Iterative Node Expansion and Path Refinement and Collision Detection and Avoidance, laying the foundation for the subsequent extension to 3D.
\subsubsection{Iterative Node Expansion and Path Refinement}

\begin{figure}[t]
\centering
\includegraphics[width=3.2in]{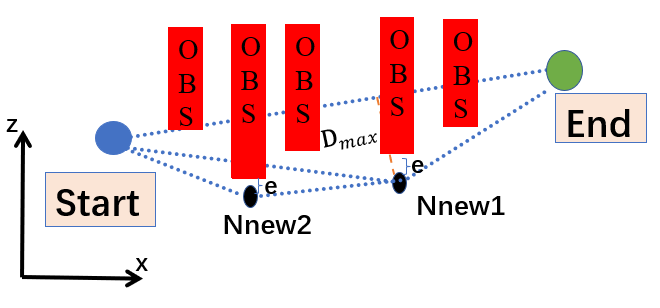}
\caption{Schematic diagram of the process of expanding path nodes.}
\label{fig5}
\end{figure}

The ILMSA algorithm generated new nodes and iteratively extended the path as follows. As shown in Fig. \ref{fig5}, the planning space initialized the starting coordinates as \(X_{\text{start}}(x_1, z_1)\) and the endpoint coordinates as \(X_{\text{end}}(x_2, z_2)\). First, a straight line was drawn between the start and end points to construct an initial path. Next, collision detection was performed on the path segment to check for any obstacles. If a collision was detected, the segment was marked, and its start and end points were recorded. The obstacle vertices, \(X_{\text{obstacle}}(x_o, z_o)\), between the start and end points were then identified. Based on Eq. (\ref{eq1}), the vertex with the maximum distance from the path segment was selected, and a key node \(N_{\text{new1}}\) was generated by offsetting the vertex by a safe distance \(e\). This process was iteratively repeated, with new nodes added to avoid obstacles, ultimately producing a collision-free path. The method ensured that each iteration reduced the likelihood of collisions, gradually refining the path until it became safe and feasible. In the example environment, a collision-free path was successfully generated after two iterations.

\begin{equation}
\label{eq1}
D_{\text{max}} = \frac{|(z_2 - z_1)x_o - (x_2 - x_1)z_o + x_2z_1 - z_2x_1|}
{\sqrt{(z_2 - z_1)^2 + (x_2 - x_1)^2}}
\end{equation}

Based on the node expansion process, we developed an algorithm for generating paths, as detailed in Algorithm \ref{alg1}. In this algorithm, the \textit{CollisionDetected} function determined whether a path segment intersected with obstacles, and the \textit{Collision Avoiding} function identified potential vertices for path refinement. If such vertices were identified, the \textit{MaxDistance} function selected the vertex furthest from the path segment, while \textit{AddNewNode} added the new node to the path. The \textit{Sort} function then ensured the nodes were ordered based on the direction from $X_{start}$ to $X_{end}$. The algorithm iterated through this process until no further collisions were detected or the maximum iteration limit was reached.

\begin{algorithm}[t]
\caption{GeneratePath ($X_{start}, X_{end}$, Obstacles)}
\label{alg1}
\begin{algorithmic}
\STATE \textbf{Input:} $X_{start}, X_{end}, Obstacles$
\STATE \textbf{Output:} Path

1. $Path_0 \gets [X_{start}, X_{end}]$

2. $dir \gets \textsc{DetermineDirection}(X_{start}, X_{end})$

3. \textbf{for} $i = 1$ \textbf{to} max\_iter \textbf{do}

4. \hspace{0.5cm} $collision\_found \gets \textbf{False}$

5. \hspace{0.5cm} \textbf{for each} segment $(s, e)$ \textbf{in} $Path$ \textbf{do}

6. \hspace{1cm} \textbf{if} \textsc{CollisionDetected}$(s, e, Obstacles)$ \textbf{then}

7. \hspace{1.5cm} $collision\_found \gets \textbf{True}$

8. \hspace{1.5cm} $V \gets \textsc{COLLISIONAVOIDING(}(s, e, Obs)$

9. \hspace{1.5cm} \textbf{if} $V \neq \emptyset$ \textbf{then}

10. \hspace{2cm} $v_{max} \gets \textsc{MaxDistance}(V, s, e)$

11. \hspace{2cm} $new\_node \gets \textsc{AddNewNode}(v_{max})$

12. \hspace{2cm} $Path \gets Path \cup \{new\_node\}$

13. \hspace{2cm} $Path \gets \textsc{Sort}(Path, dir)$

14. \hspace{1.5cm} \textbf{end if}

15. \hspace{1cm} \textbf{end if}

16. \hspace{0.5cm} \textbf{end for}

17. \hspace{0.5cm} \textbf{if} $collision\_found = \textbf{False}$ \textbf{then break}

18. \textbf{end for}

19. \textbf{return} $Path$
\end{algorithmic}
\end{algorithm}

\subsubsection{Collision Detection and Avoiding}

\begin{figure}[t]
\centering
\includegraphics[width=2.5
in]{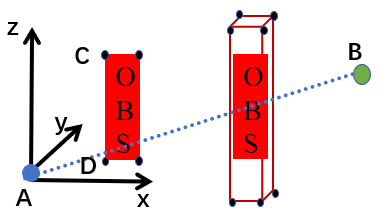}
\caption{Schematic diagram of obstacle collision detection.}
\label{fig6}
\end{figure}

ILMSA performed collision detection at each iteration. This was accomplished using a geometric method to determine whether line segments intersect, based on the counter-clockwise (CCW) orientation of three points. Eq. (\ref{eq2}) calculated whether three points \(A\), \(B\), and \(C\) were arranged in a counter-clockwise direction. As shown in Fig. \ref{fig6}, each obstacle was defined by multiple vertices forming a polygon. The algorithm iterated over all obstacles, treating each edge as a line segment. To check if the path segment \(AB\) intersected with an obstacle edge \(CD\), the conditions in Eq. (\ref{eq3}) were evaluated: if both conditions held, the segments intersected. If the path intersected any obstacle edge, a collision was detected; otherwise, the path was considered collision-free. The algorithm for collision avoiding is provided in Algorithm \ref{alg2}.

\begin{equation}\label{eq2}
\begin{split}
\text{CCW}(A, B, C) = &\ (C_Z - A_Z) \times (B_x - A_x) \\
                     > &\  (B_Z - A_Z) \times (C_x - A_x)
\end{split}
\end{equation}

\begin{equation}
\label{eq3}
\begin{split}
\text{CCW}(A, C, D) &\neq \text{CCW}(B, C, D) \quad \text{and} \\
\text{CCW}(A, B, C) &\neq \text{CCW}(A, B, D)
\end{split}
\end{equation}

\begin{algorithm}[t]
\caption{Collision Avoiding ($s, e, Obs$)}
\label{alg2}
\begin{algorithmic}
\STATE \textbf{Input:} $s, e, Obs$
\STATE \textbf{Output:} $V$

1. $V \gets \emptyset$

2. $(x_{\text{min}}, x_{\text{max}}) \gets \textsc{MinMax}(s[0], e[0])$

3. \textbf{for each} $O$ \textbf{in} $Obs$ \textbf{do}

4. \hspace{0.5cm} $v_{\text{min\_z}} \gets \textsc{MinZ}(O)$

5. \hspace{0.5cm} $V_{\text{min\_z}} \gets \{v \mid v \in O, v[1] = v_{\text{min\_z}}\}$

6. \hspace{0.5cm} \textbf{for each} $v$ \textbf{in} $V_{\text{min\_z}}$ \textbf{do}

7. \hspace{1cm} \textbf{if} $x_{\text{min}} \leq v[0] \leq x_{\text{max}}$ \textbf{and} \textsc{BelowLine}$(v, s, e)$ \textbf{then}

8. \hspace{1.5cm} $V \gets V \cup \{v\}$

9. \hspace{1cm} \textbf{end if}

10. \hspace{0.5cm} \textbf{end for}

11. \hspace{0.5cm} \textbf{for each} edge \textbf{in} $O$ \textbf{do}

12. \hspace{1cm} \textbf{if} \textsc{SegIntersect}$(s, e, \text{edge})$ \textbf{then}

13. \hspace{1.5cm} \textbf{return} \textbf{True}, $V$

14. \hspace{1cm} \textbf{end if}

15. \hspace{0.5cm} \textbf{end for}

16. \textbf{end for}

17. \textbf{return} \textbf{False}, $V$
\end{algorithmic}
\end{algorithm}

Among these functions, the \textit{MinMax} function identified the $x$-coordinate range between the start point $s$ and end point $e$, defining the area to check for obstacle vertices. The \textit{MinZ} function found the obstacle vertex with the lowest $z$-coordinate. The \textit{BelowLine} function checked whether the vertex was below the path segment, confirming its relevance for collision detection. The \textit{SegIntersect} function determined if the path segment intersected any obstacle edges, indicating a collision. If a collision occurred, the path was adjusted by selecting an avoidance vertex.

\subsection{ILMSA in 3D}
In this section, we extend ILMSA to 3D, enabling it to handle more complex spatial environments, such as those encountered in table-top strawberry harvesting.
\subsubsection{Projection of Spatial Obstacles Onto a Plane}
The aforementioned 2D environment is the vertical plane 1 as shown in Fig. \ref{fig5}. To achieve spatial path planning, we adopted the following steps.We first determined the starting point \((x_1, y_1, z_1)\) and the endpoint \((x_2, y_2, z_2)\). We then calculated the direction vector between the two points and normalized it according to formulas \ref{eq4} and \ref{eq5} to obtain the column vector \(\text{rotation\_axis}\), where \(u_x\), \(u_y\), and \(u_z\) are its three components. Next, we converted the given rotation angle to radians and chose an initial normal vector that was perpendicular to the direction vector. Using the rotation matrix from Eq. (\ref{eq8}), we rotated the initial normal vector around the rotation axis to obtain the rotated normal vector. Here, \( c = \cos \theta \), \( c' = 1 - \cos \theta \), and \( s = \sin \theta \) are the trigonometric terms used in the rotation matrix. Finally, the coefficients \(A\), \(B\), \(C\), and \(D\) of the plane equation was determined based on the rotated normal vector and the coordinates of either the starting or endpoint, thereby generating a plane passing through the two points with a specific rotation angle for path planning.

\begin{equation}
\label{eq4}
\text{direction\_vector} = \begin{pmatrix} x_2 - x_1 \\ y_2 - y_1 \\ z_2 - z_1 \end{pmatrix}
\end{equation}


\begin{equation}
\setlength{\belowdisplayskip}{15pt}  
\label{eq5}
\text{rotation\_axis} = \frac{\text{direction\_vector}}{\sqrt{(x_2 - x_1)^2 + (y_2 - y_1)^2 + (z_2 - z_1)^2}}
\end{equation}

\begin{equation}
\label{eq8}
R = \begin{bmatrix}
c + u_x^2 c' & u_x u_y c' - u_z s & u_x u_z c' + u_y s \\
u_y u_x c' + u_z s & c + u_y^2 c' & u_y u_z c' - u_x s \\
u_z u_x c' - u_y s & u_z u_y c' + u_x s & c + u_z^2 c'
\end{bmatrix}
\end{equation}

Next, each spatial point in the obstacle set was projected onto the plane. For each obstacle, which consisted of multiple 3D points, the projection function was applied to map each point onto the plane, with the results stored in a set of projected obstacles. This process produced a set of coordinates representing the projection of all 3D obstacle points onto the plane, ensuring accurate mapping for further processing and analysis. The detailed steps of the projection algorithm are presented in Algorithm \ref{alg3}.

\begin{algorithm}[t]  
\caption{Projection on Plane}
\label{alg3}
\begin{algorithmic}
\STATE \textbf{Input:} $A, B, C, D, p$ \textbf{or} $A, B, C, D, \text{Obs}$
\STATE \textbf{Output:} $p'$ \textbf{or} $\text{projected\_obstacles}$

\STATE 1. \textsc{Projection}$(A, B, C, D, p)$

\STATE 2. $t \gets \frac{A \cdot p.x + B \cdot p.y + C \cdot p.z + D}{A^2 + B^2 + C^2}$

\STATE 3. $p' \gets (p.x - A \cdot t,\, p.y - B \cdot t,\, p.z - C \cdot t)$

\STATE 4. \textbf{return} $p'$

\STATE 5. \textsc{ProjectObstaclesOnPlane}$(A, B, C, D, \text{Obs})$

\STATE 6. \textbf{for each} $O$ \textbf{in} $\text{Obs}$ \textbf{do}

\STATE 7. \hspace{0.5cm} $P \gets \{\textsc{Projection}(A, B, C, D, p) \mid p \in O\}$

\STATE 8. \hspace{0.5cm} Append $P$ to $\text{projected\_obstacles}$

\STATE 9. \textbf{end for}

\STATE 10. \textbf{return} $\text{projected\_obstacles}$
\end{algorithmic}
\end{algorithm}

\subsubsection{Spatial Path Optimization.}

To find the optimal spatial path, after identifying a path on the projection plane, the plane’s rotation angle was altered to generate a series of additional planes, and the path search was repeated on each. Since the initial path may contain abrupt directional changes, B-spline curves were employed for smoothing. The specific steps is outlined in Algorithm \ref{alg4}, which constructed a smooth 3D B-spline curve by computing the coordinates of curve points using the \textit{de\_Boor} function. The process began by initializing the \textit{KnotVector} $T$, which divided the curve into segments. For each segment, the control points were used to calculate the coordinates $(x, y, z)$ via the recursive \textit{de\_Boor} algorithm. These computed points were then appended to the set of data points. The final output was a list of 3D coordinates representing the smooth B-spline curve.

\begin{algorithm}[t] 
\caption{Generate 3D B-Spline Curve}
\label{alg4}
\begin{algorithmic}
\STATE \textbf{Input:} $\text{control\_points}$
\STATE \textbf{Output:} $data\_points$

\STATE 1. \textsc{GenerateBSpline3D}$(\text{control\_points})$

\STATE 2. \hspace{0.5cm} Initialize: $T \gets \text{KnotVector}$, $data\_points \gets []$

\STATE 3. \hspace{0.5cm} \textbf{For each} segment $[T_j, T_{j+1}]$:

\STATE 4. \hspace{1cm} \textbf{Compute} $x, y, z \gets \text{de\_Boor}$

\STATE 5. \hspace{1cm} \textbf{Append} $(x, y, z)$ to $data\_points$

\STATE 6. \hspace{0.5cm} \textbf{Return} $data\_points$
\end{algorithmic}
\end{algorithm}

Within the planning space, we generated multiple collision-free paths. To find the optimal path, a path quality evaluation function was established. We comprehensively considered path length, safety, and smoothness. Path safety was determined by calculating the minimum distance from each point on the path to the nearest obstacle edge. Path smoothness was calculated by summing the angles between consecutive path segments, which were determined by the vectors formed between adjacent points along the path. The comprehensive score combined these three metrics using weighted values, with \(w_{\text{length}}\), \(w_{\text{safety}}\), and \(w_{\text{smoothness}}\), respectively.
The final score reflected the overall quality of the path, with a lower score indicating better path quality.

\subsection{Algorithm Implementation Process}

\begin{figure}[!t]
\centering
\includegraphics[width=3
in]{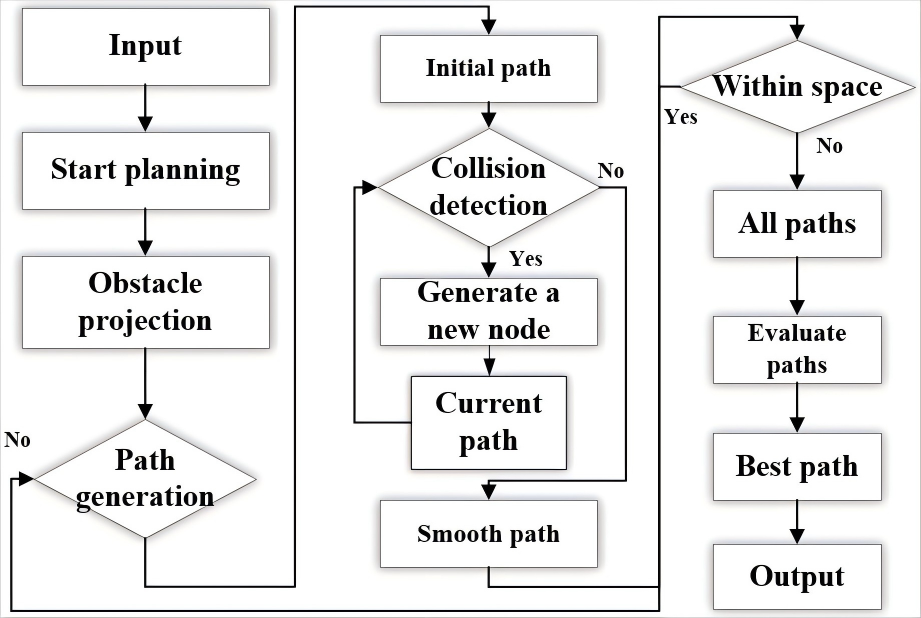}
\caption{Flow chart of the algorithm.}
\label{fig7}
\end{figure}
\begin{figure}[!t]
\centering
\includegraphics[width=3
in]{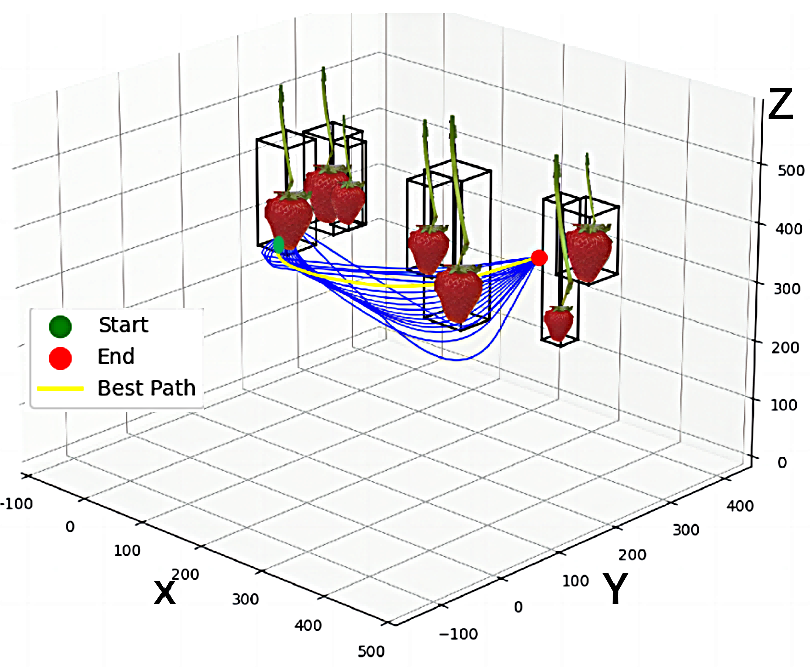}
\caption{Planning effect of ILMSA algorithm in table-top grown strawberries environment.}
\label{fig8}
\end{figure}

The entire path-planning algorithm operated as an independent ROS node, integrating the various sub-modules described above. As shown in Fig. \ref{fig7}, the flowchart illustrates the algorithm's process. Upon receiving the start and end points, the program initiated after the obstacle information was perceived. The obstacle projection module transferred the path planning to a spatial plane, where collision detection was performed and path nodes were iteratively generated. The algorithm rotated the projection plane in 5 degree increments, exploring paths until the entire planning space was covered. Afterward, the algorithm smoothed the paths and conducted a quality assessment. The path with the lowest score was selected as the optimal path. Fig. \ref{fig8} shows path planning results in the table-top grown strawberries environment, where ILMSA generated multiple collision-free spatial paths, and after a quality assessment, the yellow path was determined as the best one.

\section{Experiments}
To validate the new algorithm, we first conducted simulation experiments.  In the 3D environment, ILMSA was compared with the 3D-RRT and LPS algorithms, while in the 2D environment, it was compared with the RRT, RRT-Connect and A* algorithms, as well as with the learning-based algorithm QAPF. Later, the algorithm was deployed on a harvesting robot to verify its effectiveness in continuous path planning for fruit picking in real-world scenarios. The simulations were implemented using Python 3.9, with the hardware platform on a Windows 11 operating system equipped with an i7-12650H 2.30 GHz CPU.

\subsection{Simulation experiment in simple environment}

\begin{figure}[!t]
\centering
\includegraphics[width=3.5in]{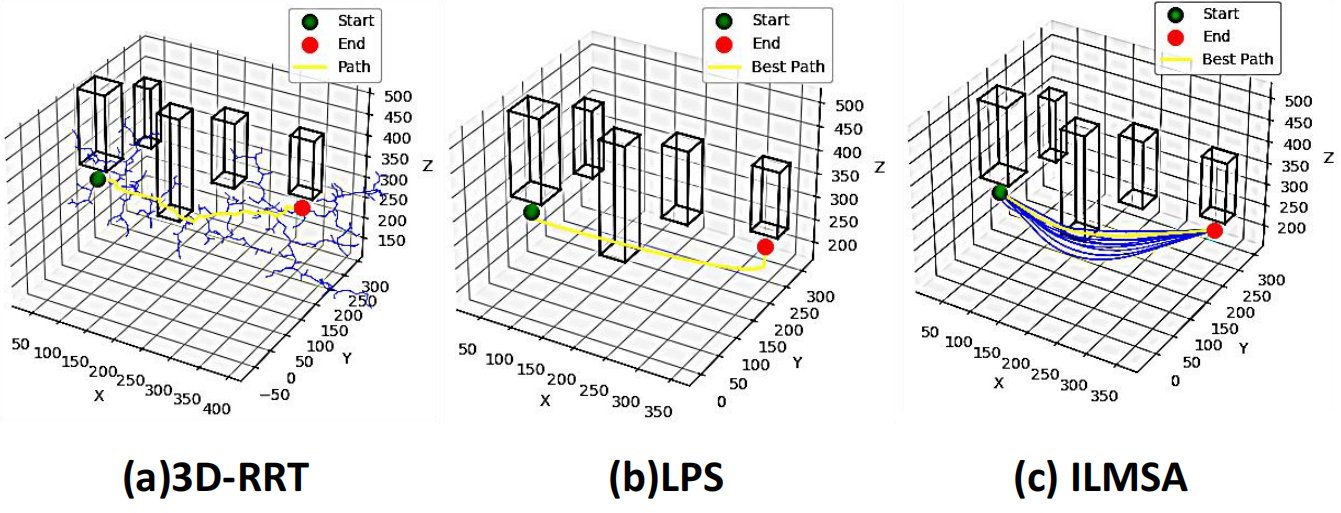}
\caption{Performance of different 3D path-planning algorithms in simple environment.}
\label{fig9}
\end{figure}
\begin{figure}[!t]
\centering
\includegraphics[width=3.4in]{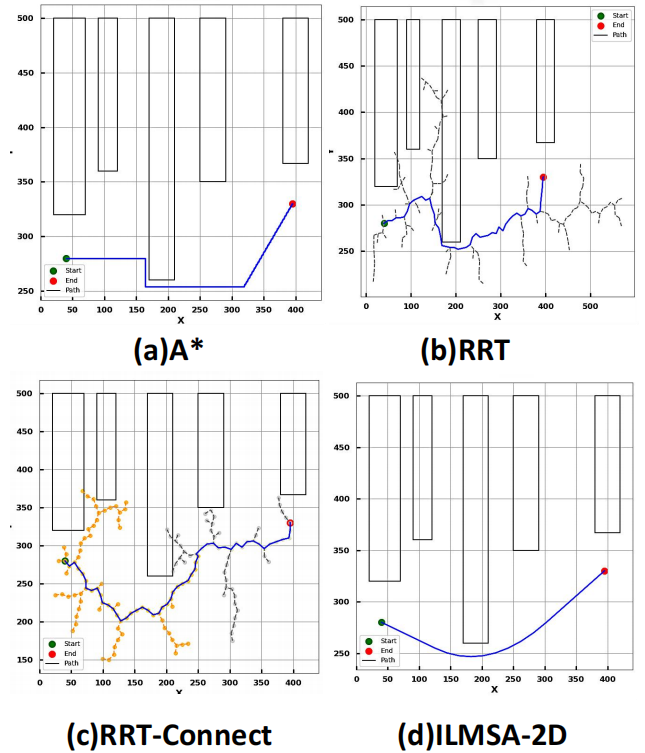}
\caption{Performance of different 2D path-planning algorithms in simple environment.}
\label{fig10}
\end{figure}

To evaluate the performance of the algorithm in environments with varying complexity, we conducted experiments in several different strawberry-harvesting scenarios. First, in the simple environment (Environment 1), we ran simulations for five strawberries to be harvested. The 3D simulation space ranged from x: 0–400 mm, y: 0–300 mm, and z: 0–500 mm, with start coordinates (40 mm, 120 mm, 280 mm) and end coordinates (395 mm, 145 mm, 330 mm). Given the randomness of sampling-based path-planning algorithms, each algorithm was tested 50 times. Fig. \ref{fig9} shows the results of three algorithms in Environment 1. In Fig. \ref{fig9}(a), the 3D-RRT algorithm took 1.01 seconds, with 208 path nodes and a final path length of 482.21 mm. In Fig. \ref{fig9}(b), the LPS algorithm took 0.019 seconds, with 201 nodes and a path length of 413.45 mm. The last image in Fig. \ref{fig9} shows the performance of the ILMSA, which achieved a total planning time of 0.035 seconds, with only 151 nodes and a reduced path length of 378.24 mm. These results demonstrated that ILMSA significantly reduced path length and redundant nodes compared to conventional sampling algorithms, offering smoother paths and better planning efficiency in high-dimensional spaces. In 2D environments, we projected all points from Environment 1 onto the xoz plane for path planning. As shown in Fig. \ref{fig10}, we compared A*, RRT, RRT-Connect, and ILMSA. The results showed that ILMSA produced much smoother paths with shorter path lengths. 

\begin{figure*}[!t]
\centering
\subfloat[]{\includegraphics[width=2.35in]{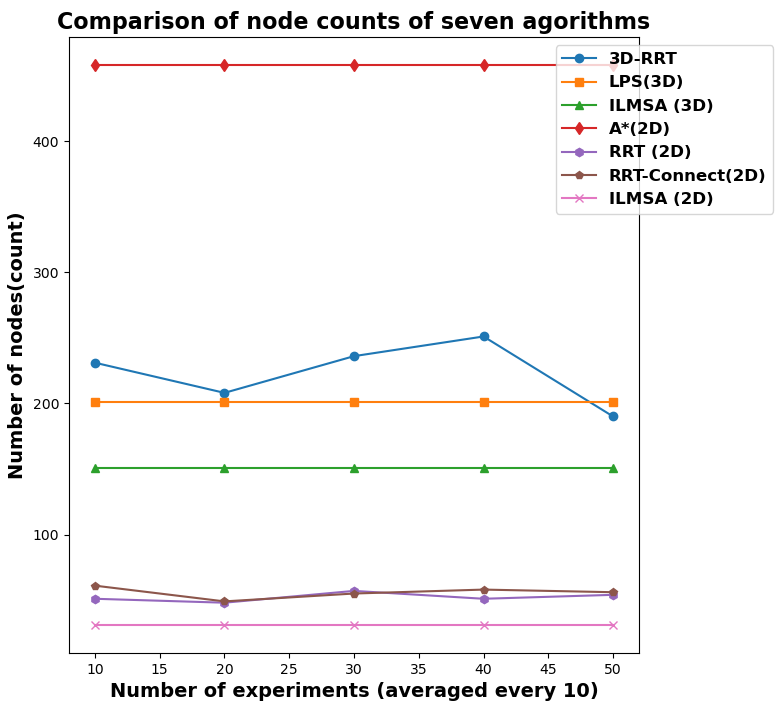}%
\label{fig_case1}}
\hfil
\subfloat[]{\includegraphics[width=2.3in]{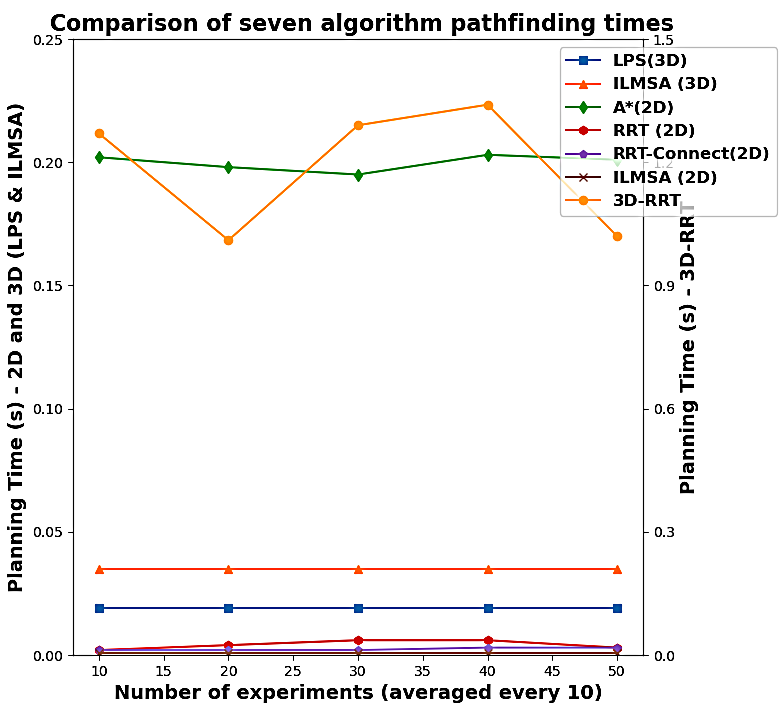}%
\label{fig_case2}}
\hfil
\subfloat[]{\includegraphics[width=2.3in]{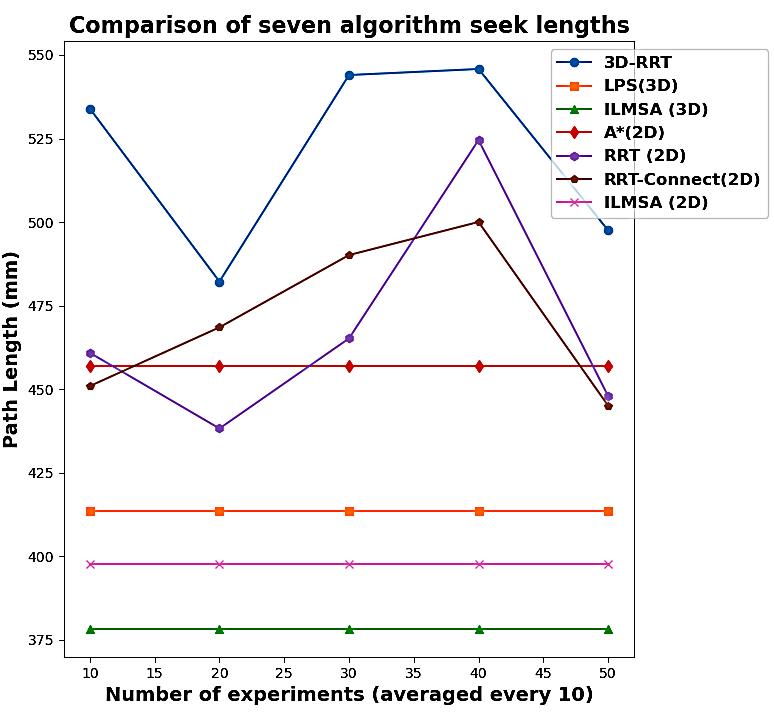}%
\label{fig_case3}}
\caption{Data Analysis of Seven Algorithms in Simple Environment: (a) node count of different algorithms under multiple experiments, (b) planning time of different algorithms under multiple experiments, (c) path length of different algorithms under multiple experiments.}
\label{fig11}
\end{figure*}

To validate the stability of the algorithm, the performance of the seven algorithms was repeated 10, 20, 30, 40, and 50 trials in Environment 1. Three key performance metrics were compared: node count, planning time, and path length, to evaluate the efficiency and quality of the algorithms. As shown in Fig. \ref{fig11}(a) and \ref{fig11}(b), we analyzed the relationship between node count and search time for the seven algorithms in Environment 1. Through effective node expansion, node count was significantly reduced in both 2D and 3D environments, enabling faster search speeds. Fig. \ref{fig11}(c) compares the path lengths of the seven algorithms in Environment 1. After both a small number of trials (10) and a large number of trials (50), ILMSA consistently found the shortest path in both 2D and 3D spaces. The results from Fig. \ref{fig11} demonstrate that ILMSA not only adapted well to both low- and high-dimensional environments, but also completed path planning more quickly and efficiently.

\subsection{Simulation experiment in complex environment}

To evaluate the performance of the algorithm in complex environments, we conducted an experiment in Environment 2, which included 13 strawberries to be harvested. The 3D simulation space ranged from x: 0–500 mm, y: 0–300 mm, and z: 0–500 mm, with start coordinates at (40 mm, 120 mm, 280 mm) and end coordinates at (465 mm, 145 mm, 330 mm). In the 2D environment, all points were projected onto the xoz plane for path planning, with start coordinates at (40 mm, 280 mm) and end coordinates at (465 mm, 330 mm). Each algorithm was tested 50 times, and the average results are shown in Table \ref{label2}. To assess the performance differences among the algorithms in a complex environment, we conducted Kruskal-Wallis tests on key metrics such as planning time, path length, and node count. The tests indicated significant differences across all metrics (\(p < 0.001\)), confirming that the algorithms performed differently.
Post-hoc pairwise comparisons were then performed using the Mann-Whitney U test, with results summarized in Table~\ref{tab_statistical_analysis}. The significance level for all tests was set to 0.05.


\begin{table}[!t]
\centering
\caption{Simulation results of seven algorithms in complex environment}
\label{label2}
\begin{tabular}{p{1cm}cccc}
\hline
\textbf{Test environment} & \textbf{Algorithm} & \textbf{nodes(count)} & \textbf{time(s)} & \textbf{length(mm)} \\
\hline
            & 3D-RRT     & 268  & 1.36  & 606.88 \\
\textbf{3D} & LPS        & 201  & 0.01  & 538.66 \\
            & ILMSA(ours)& 150  & 0.04  & 476    \\
\hline
            & A*         & 458  & 0.219 & 607    \\
\textbf{2D} & RRT        & 74   & 0.036 & 663.71 \\
            & RRT-Connect& 61 & 0.027  & 642.53   \\
            & ILMSA(ours)& 31   & 0.001 & 508.43 \\
\hline
\end{tabular}
\end{table}

\begin{itemize}
    \item In the 2D environment, ILMSA significantly outperformed A*, RRT, and RRT-Connect in terms of path length (\(p < 0.01\)), planning time (\(p < 0.01\)), and node count (\(p < 0.01\)).
    \item In the 3D environment, ILMSA achieved significantly shorter path lengths compared to 3D-RRT (\(p < 0.01\)), while its planning time was faster than 3D-RRT (\(p < 0.01\)) but comparable to LPS.
    \item ILMSA provides shorter paths and faster computations in both 2D and 3D environments, confirming its superiority in path planning efficiency and effectiveness.
\end{itemize}

\begin{table}[!t]
\centering
\caption{Statistical Analysis of ILMSA Performance in 3D and 2D Environments}
\label{tab_statistical_analysis}
\begin{threeparttable}
\begin{tabular}{p{2.2cm} c c c c}
\hline
\textbf{Comparison}        & \textbf{Path Length} & \textbf{Planning Time} & \textbf{Node Count} \\
\hline
ILMSA vs 3D-RRT           & 0.001 (Sig.)         & 0.002 (Sig.)          & 0.003 (Sig.)        \\
ILMSA vs LPS              & 0.12 (NS)            & 0.05 (NS)             & 0.08 (NS)           \\
ILMSA(2D) vs A*           & 0.0005 (Sig.)        & 0.002 (Sig.)          & 0.01 (Sig.)         \\
ILMSA(2D) vs RRT          & 0.0005 (Sig.)        & 0.002 (Sig.)          & 0.01 (Sig.)         \\
ILMSA(2D) vs RRT-Connect  & 0.0005 (Sig.)        & 0.002 (Sig.)          & 0.01 (Sig.)         \\
\hline
\end{tabular}
\begin{tablenotes}
\small
\item *$p < 0.05$ (Statistically Significant), NS: Not Significant
\end{tablenotes}
\end{threeparttable}
\end{table}

The statistical analysis results demonstrate ILMSA's consistent advantages over other algorithms in both 2D and 3D environments, particularly in terms of planning time and path length. However, the comparable performance of ILMSA and LPS in certain metrics, especially in the 3D environment, warrants further investigation.

\begin{figure*}[!t]
\centering
\subfloat{\includegraphics[width=6.9in]{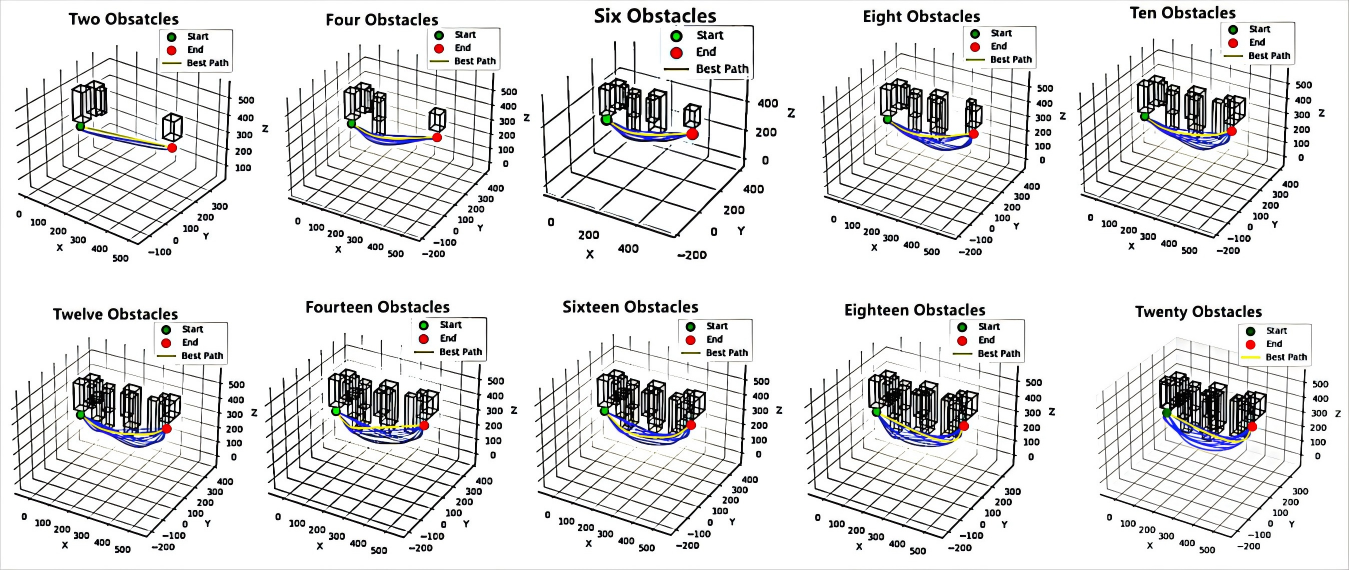}%
\label{fig_second}} 
\caption{Performance of path planning in 10 different strawberry picking environments.}
\label{fig12}
\end{figure*}

\begin{figure}[!t]
\centering
\includegraphics[width=3.3in]{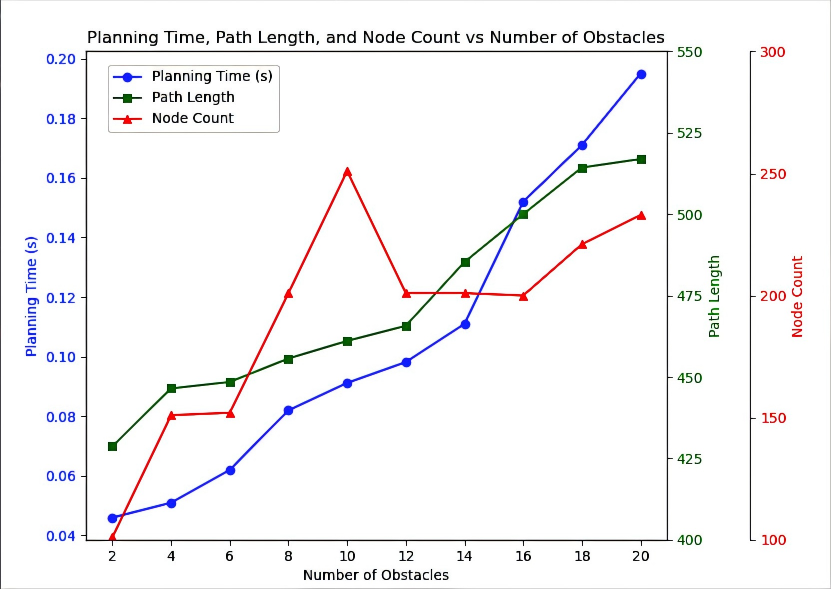}
\caption{Diagram of ILMSA's path-planning time, path length and number of path nodes changing with the number of obstacles.}
\label{fig13}
\end{figure}

To further explore the algorithm's adaptability to different environments, we incrementally increased the complexity by varying the number of obstacles from 2 to 20 in steps of 2. Start and end parameters were consistent with those in Environment 2. The planning results demonstrated that the algorithm successfully found an optimal, collision-free path in all environments, as shown in Fig. \ref{fig12}. To account for randomness, each scenario was tested 10 times, and the average planning time, path length, and number of path nodes were recorded and plotted in Fig. \ref{fig13}. As obstacle numbers increased, planning time rose gradually from approximately 0.04 to 0.20 seconds, and path length increased by around 30\%, from 425 mm to 550 mm. This indicates that greater complexity required more computation time, though still within the millisecond range, while path length grew slowly as the algorithm searches for the optimal route. The number of path nodes fluctuated from 100 to 250, reflecting the increased complexity of node expansion. Overall, the algorithm adapted well to complex obstacle environments and met the real-time millisecond-level requirements.

\subsection{Simulation experiment in specific scenarios}

\begin{figure}[!t]
\centering
\includegraphics[width=3.6in]{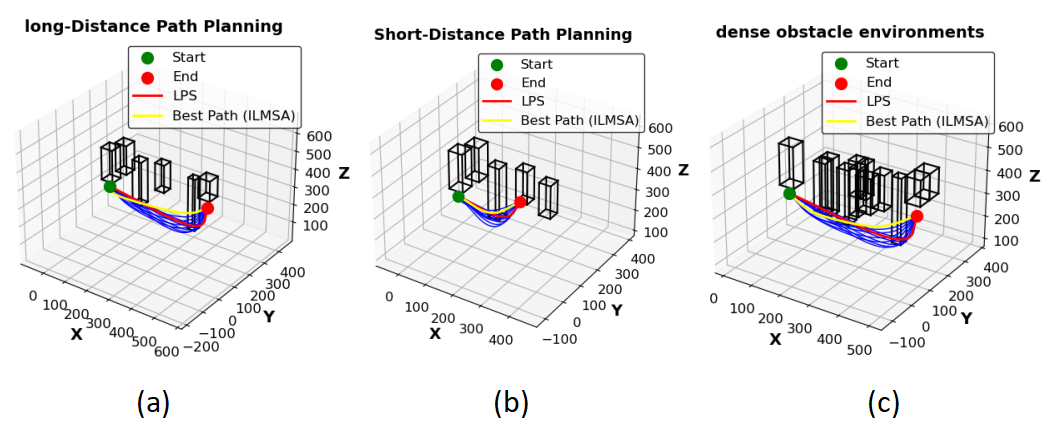}
\caption{Comparison of LPS and ILMSA in additional scenarios: (a) Long-distance path planning, (b) Short-distance path planning, and (c) Dense obstacle environments. }
\label{supplementary_experiments}
\end{figure}

To further evaluate ILMSA, additional simulation experiments were designed, focusing on specific scenarios: short-distance, long-distance, and dense obstacle environments, as shown in Fig. \ref{supplementary_experiments}. These scenarios also reflect real-world growth conditions of table-top strawberries. In addition to comparing ILMSA with the advantageous LPS algorithm, we also included a comparison with the learning-based algorithm QAPF to further demonstrate the superiority of the proposed method.

When comparing with LPS, each scenario was simulated 50 times, with performance metrics including path length, planning time, and number of path nodes. The average values of these metrics are presented in Table \ref{tab_detailed_comparison}. Statistical analyses were conducted using the Mann-Whitney U test, which confirmed that ILMSA significantly outperforms LPS in terms of node count and path length (\(p < 0.01\)), while there was no significant difference in planning time (\(p > 0.05\)). These results demonstrate that ILMSA is more effective than LPS in finding shorter paths and reducing redundant nodes, while achieving comparable planning time performance.

\begin{table}[!t]
\centering
\caption{Performance Comparison of ILMSA and LPS in Different Scenarios.}
\label{tab_detailed_comparison}
\begin{tabular}{lccc}
\hline
\textbf{Scenario} & \textbf{Metric} & \textbf{ILMSA} & \textbf{LPS} \\
\hline
                        & Planning Time (s) & 0.02 & 0.01 \\
Short Distance & Path Length (mm)   & 120  & 135  \\
               & Nodes (count)      & 20   & 28   \\
\hline
                 & Planning Time (s) & 0.08 & 0.7 \\
  Long Distance              & Path Length (mm)   & 320  & 375  \\
               & Nodes (count)      & 70   & 85   \\
\hline
                & Planning Time (s) & 0.12 & 0.09 \\
  Dense Obstacles    & Path Length (mm)  & 250  & 290  \\
                & Nodes (count)     & 60   & 78   \\
\hline
\end{tabular}
\end{table}

\begin{figure}[!t]
\centering
\includegraphics[width=3.6in]{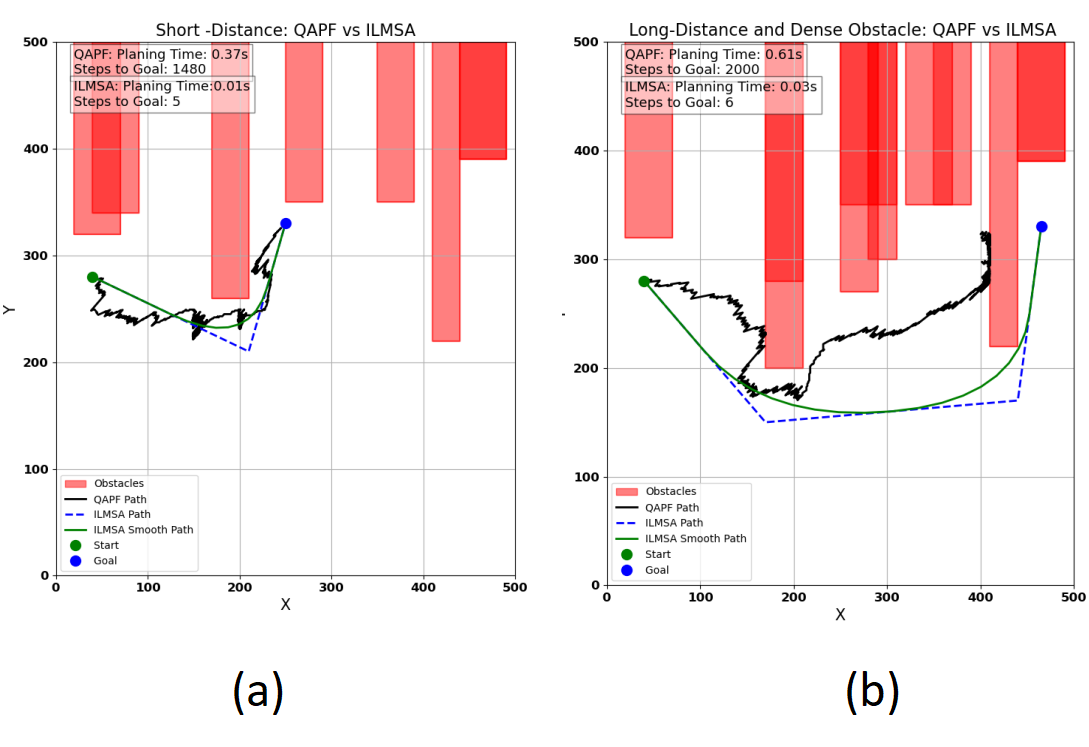}
\caption{Comparison of QAPF and ILMSA in additional scenarios: (a) Short-distance (d) Complex scenario:Long-distance and dense obstacle.}
\label{QAPF}
\end{figure}

When comparing QAPF with ILMSA, we projected all points from specific scenarios onto the xoz plane, analogous to a 2D map used in mobile robot path planning. Although QAPF, as a representative advanced path planning algorithm, has demonstrated good performance in mobile robot applications, the results of our comparative experiments were suboptimal. To avoid experimental redundancy, we integrated long-distance and dense obstacle as complex scenario. Each scenario was also simulated 50 times, with performance metrics including path length, planning time, and planning success rate. The average values of these metrics are presented in Table \ref{QAPF_ILMSA_comparison} and the test results was shown in Fig. \ref{QAPF}. In short-distance scenario, QAPF achieved a path planning success rate of 85\% compared to 100\% of ILMSA. 
The paths generated by QAPF were excessively rugged. Compared to QAPF, ILMSA reduced path length by 36.7\% and shortened planning time by 97.8\%.
In complex scenario, QAPF consistently became trapped in local oscillations despite the guidance provided by Q-learning for action decision-making. In contrast, ILMSA employed iterative node expansion and path refinement methods across all scenarios, rapidly identifying obstacle-avoiding nodes and generating smooth paths through post-processing. We carefully considered the reasons for the observed performance differences: first, unlike the 2D maps used for mobile robots, our 2D table-top strawberry environments impose more constraints and are narrower; second, QAPF involves numerous parameters, such as learning rate, discount factor, attractive gain, and repulsive gain, where different configurations significantly impact the results. Moreover, increased training data leads to a substantial increase in planning time. These findings demonstrate that ILMSA not only overcomes the limitations of QAPF in complex and constrained environments but also achieves more efficient and smoother path planning, making it more suitable for applications in continuous harvesting of strawberries.

\begin{table}[!t]
\centering
\caption{Performance Comparison of ILMSA and QAPF in 2D Scenarios.}
\label{QAPF_ILMSA_comparison}
\begin{tabular}{lccc}
\hline
\textbf{Scenario} & \textbf{Metric} & \textbf{ILMSA} & \textbf{QAPF} \\
\hline
             & Planning Time (s) & 0.01 & 0.45 \\
Short Distance               & Path Length (mm)   & 160  & 253  \\
               & planning success rate (\%)     & 100   & 85   \\
\hline
                    & Planning Time (s) & 0.03 & N/A \\
 Complex scenario              & Path Length (mm)   & 510  & N/A  \\
               & planning success rate(\%)     & 98   & N/A   \\

\hline
\end{tabular}
\end{table}

\subsection{Field Test}

\begin{figure}[!t]
\centering
\includegraphics[width=2 in]{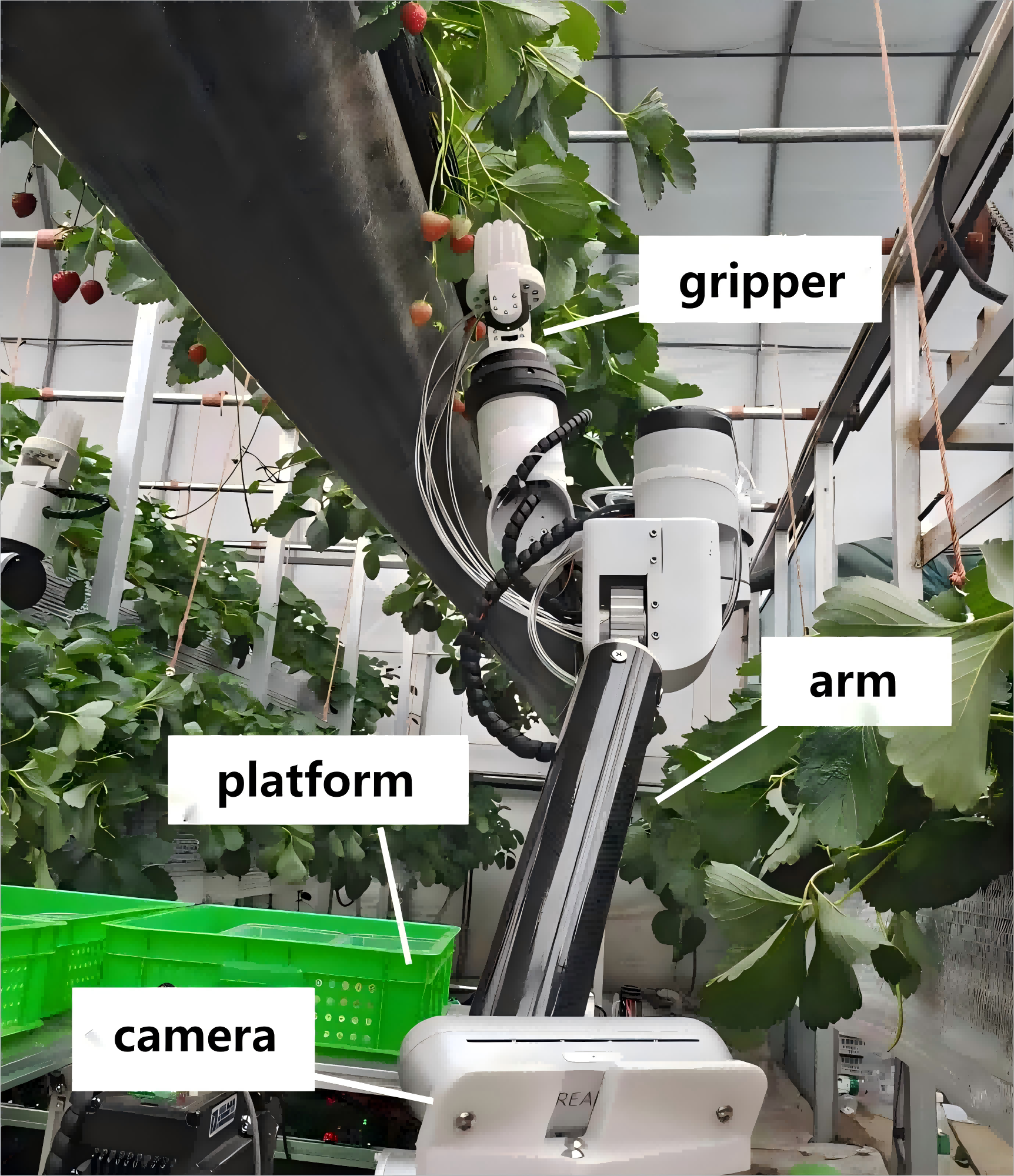}
\caption{The strawberry harvesting robot used in this study.}
\label{fig14}
\end{figure}

To test the path-planning capability of the proposed algorithm in a real-world strawberry-harvesting environment, we deployed the algorithm to a strawberry-harvesting robot running on Ubuntu 20.04. The experiments were conducted at Shennong Tiandi Strawberry Farms, located in Shunyi District, Beijing. As shown in Fig. \ref{fig14}, the robot used in the experiment was composed of a newly developed hybrid 6-DoF robotic arm, a Realsense D455 camera, and a computer equipped with the ROS system \cite{chen2024design}.

\begin{figure}[!t]
\centering
\includegraphics[width=3.4
in]{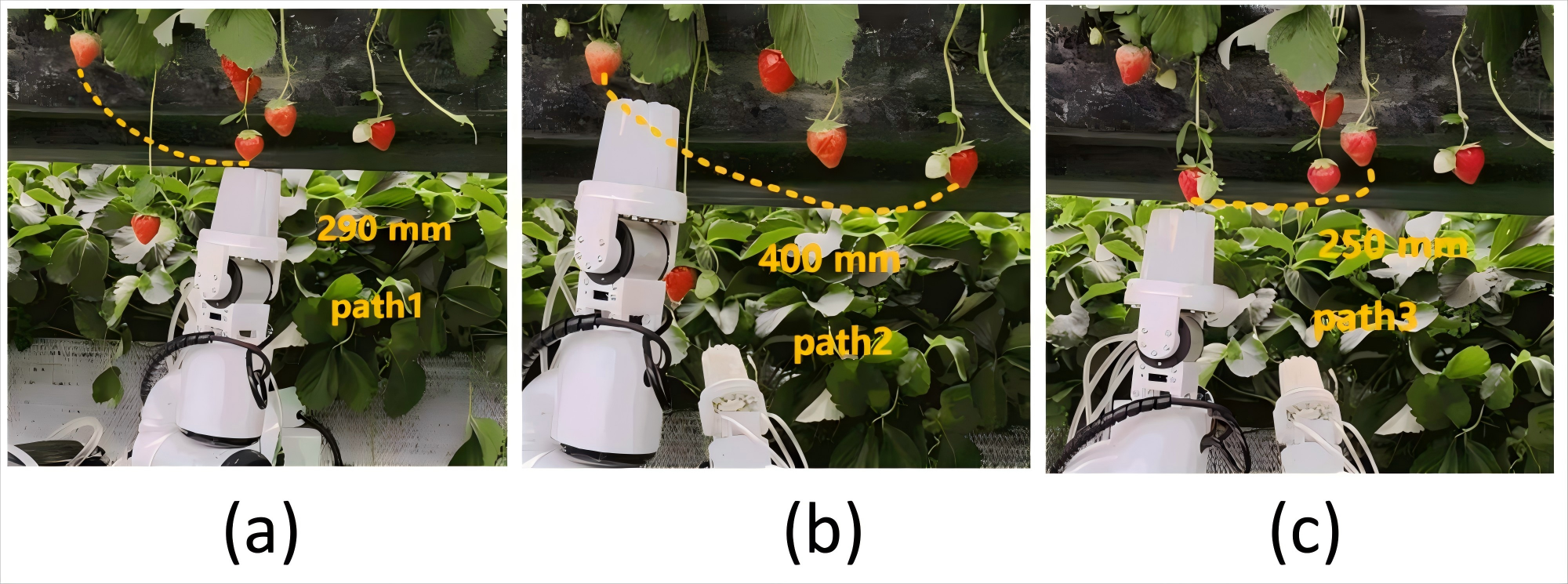}
\caption{Three common strawberry distribution scenarios: the yellow line represents path for continuous harvesting, avoiding obstacles from the side.}
\label{fig15}
\end{figure}

\begin{figure*}[!t]
\centering
\subfloat{\includegraphics[width=6.8in]{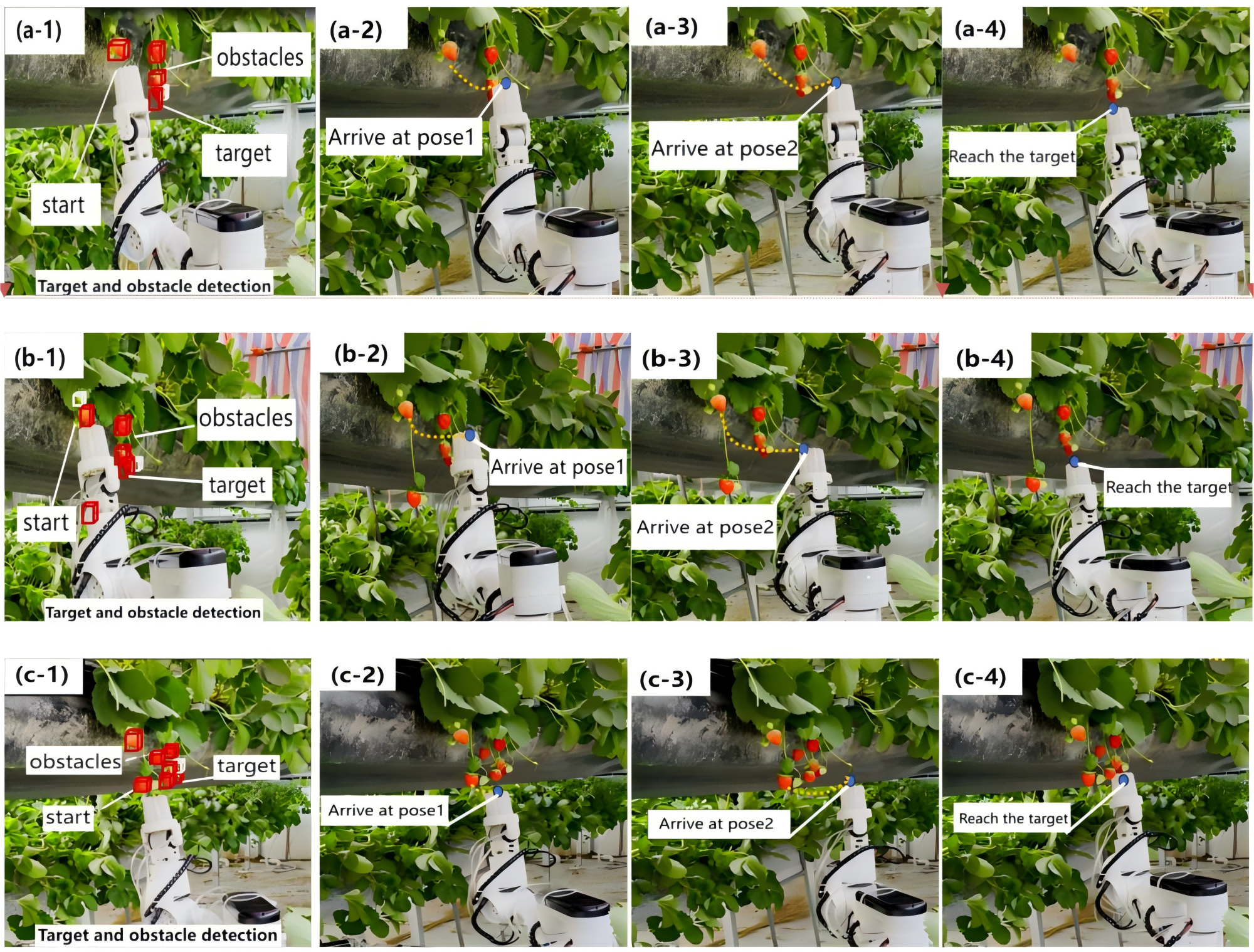}%
\label{fig_second_case}} 
\caption{Path-planning process in three typical scenarios: (a) short-distance detours, (b) long-distance detours, (c) clustered strawberries.}
\label{fig16}
\end{figure*}

The field testing process was as follows: after the robotic arm reached its initial posture, the D455 camera began target detection; once detection was complete, the positional information of the nearest ripe strawberry was sent to the robotic arm for harvesting. We selected three common strawberry distribution scenarios: (a) short-distance detours, (b) long-distance detours, and (c) clustered strawberries, with the planned trajectories shown in Fig. \ref{fig15}. However, in the experiment, we focused solely on the harvesting paths, so the picking hand was only directed toward the strawberries without actually picking them. To better visualize the trajectories, we plotted them in Fig. \ref{fig16}. The trajectory tracking time for the picking hand was 0.01 seconds as it navigated through all nodes, avoiding unpicked strawberries and reaching the target strawberry before initiating the next picking task. The average execution times across multiple planning runs for each scenario were 1.6 seconds, 2.0 seconds, and 1.3 seconds, respectively. These results validate ILMSA's ability to perform real-time, collision-free path planning for multiple strawberries in real-world conditions.

\begin{figure}[!t]
\centering
\includegraphics[width=3.4in]{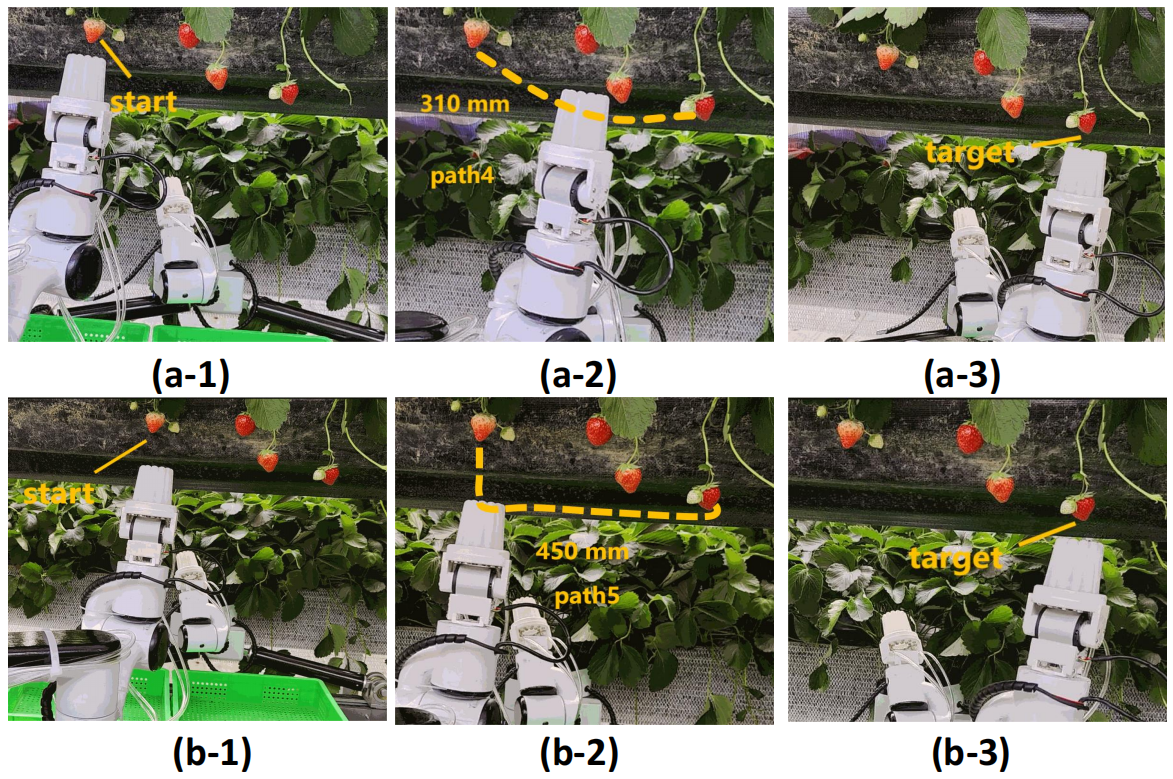}
\caption{Comparison of continuous path planning between new algorithm and LPS algorithm: (a) path-planning performance of ILMSA algorithm, (b) path-planning performance of LPS strategy. }
\label{fig17}
\end{figure}

\begin{figure}[!t]
\centering
\includegraphics[width=3.3in]{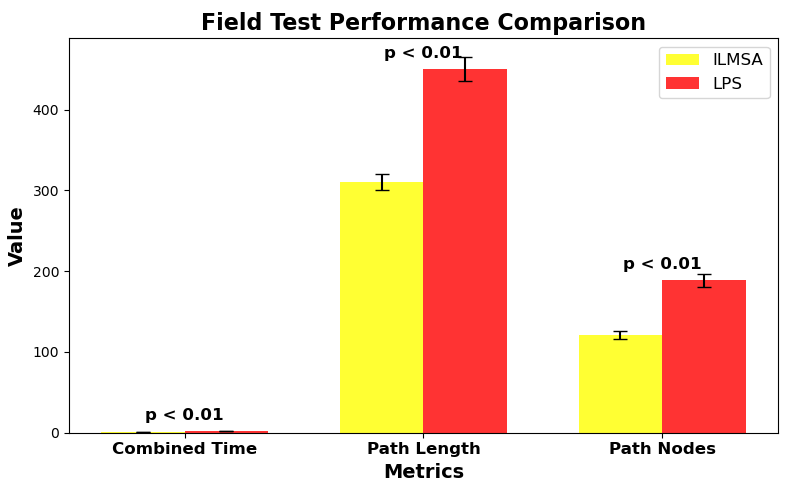}
\caption{Statistical comparison of ILMSA and LPS in the field test. Metrics include combined planning and execution time, path length, and number of path nodes, with error bars representing standard deviations.}
\label{fig:field_test_stats}
\end{figure}

To further evaluate the performance of the ILMSA algorithm, we compared it with the heuristic LPS method. In the experiment, both path-planning algorithms were deployed on the robot, guiding the picking hand to move back and forth between the start and target points, as shown in Fig. \ref{fig17}. Over 50 trials, we recorded the planning time, execution time, path length, and the number of path nodes. As shown in Table \ref{tab2}, ILMSA demonstrated a combined planning and execution time that is approximately 58\% of LPS's time, and its average path length is 69\% of that of LPS. Additionally, the number of path nodes generated by ILMSA is significantly lower than that of LPS, highlighting the algorithm's ability to minimize unnecessary node expansions. To validate these observed differences, a Mann-Whitney U test was performed for each metric. The results revealed significant differences between ILMSA and LPS across all metrics, with p-values below 0.01. Fig. \ref{fig:field_test_stats} visualizes the comparison of ILMSA and LPS across these metrics.

\begin{table}[t]
\centering
\caption{Comparison of ILMSA and LPS algorithms}
\label{tab2}
\begin{tabular}{lccc}
\hline
\textbf{Parameters} & \textbf{ILMSA(ours)} & \textbf{LPS} \\
\hline
Combined Time (s) & 1.4 & 2.4 \\
Path Length (mm) & 310 & 450 \\
Number of Path Nodes (pcs) & 121 & 189 \\
\hline
\end{tabular}
\end{table}

The results demonstrated that the ILMSA algorithm reliably found damage-free picking paths, avoided redundant detours, and significantly reduced execution time. Statistical analyses further confirm ILMSA's robustness and efficiency, making it a promising effective solution for real-world agricultural robotics.

\subsection{Discussion}
Based on the experiments above, our path-planning method has shown advantages over conventional, experienced, and learning-based methods in terms of planning time, path length, safety, and smoothness. Despite the successful application in table-top grown strawberries, there are several limitations that need to be addressed. First, the collision detection system simplifies strawberries into cuboid bounding models. While this abstraction is effective for the current experiments, it may not fully capture the irregular shapes of real strawberries, potentially leading to detection errors in more complex scenarios. Second, although the algorithm has few parameters that are easy to configure, its performance is still influenced by certain parameters. In our experiments, we identified the following parameters as critical for achieving optimal performance:

\begin{itemize}
    \item \textit{Collision Distance Threshold ($e$)}: {This parameter defines the safe offset distance for generating new path nodes. Smaller values (e.g., $e = 1$ mm) increase collision risks but result in shorter paths, while larger values (e.g., $e = 10$ mm) improve safety by preventing the gripper from approaching delicate stems, though they may lead to longer paths and increased planning time, especially in dense environments. A balanced setting of $e = 5$ mm was used in our experiments.}
    
    \item \textit{Rotation Angle Increments ($\Delta \theta$)}: {This parameter determines the granularity of the projection plane's rotation during spatial path searching. Finer increments (e.g., $\Delta \theta = 1^\circ$) improve the algorithm's ability to find optimal paths but increase computational overhead. Coarser increments (e.g., $\Delta \theta = 10^\circ$) reduce computation time but may result in suboptimal paths, particularly in environments with complex obstacles. We used $\Delta \theta = 5^\circ$, which provided a good trade-off between computational efficiency and path quality.}
    
    \item \textit{Path Evaluation Weights ($w_{\text{length}}, w_{\text{safety}}, w_{\text{smoothness}}$)}: {These weights define the relative importance of path length ($w_{\text{length}}$), safety ($w_{\text{safety}}$), and smoothness ($w_{\text{smoothness}}$) in the path quality evaluation. Increasing $w_{\text{smoothness}}$ emphasizes smoother paths but often increases planning time, while prioritizing $w_{\text{length}}$ minimizes path length at the expense of smoothness. A balanced configuration of $w_{\text{length}} = 0.4$, $w_{\text{safety}} = 0.4$, and $w_{\text{smoothness}} = 0.2$ was found to effectively balance path quality and computational efficiency.}
\end{itemize}

\section{conclusion}


A novel collision-free path-planning method for the continuous harvesting of table-top grown strawberries has been proposed, integrating the strengths of the artificial potential field method, graph search algorithms, and the RRT algorithm. A path node generation and expansion mechanism enables rapid node expansion and refinement in complex environments, compatible with both high- and low-dimensional spaces. The spatial obstacle projection method, combined with collision detection, effectively avoids path collisions in high-dimensional spaces. B-spline curve smoothing and a path quality evaluation function were used to identify the optimal collision-free path. ILMSA significantly reduces path lengths, planning time, and computational complexity compared to traditional algorithms such as 3D-RRT, LPS, A*, and RRT-Connect. Specifically, due to its low computational requirements, ILMSA does not necessitate extensive parameter configuration or data training, rendering it suitable for high-dimensional environments. Furthermore, ILMSA is more adept for path planning in table-top grown strawberrie compared to learning-based algorithms like QAPF. These benefits were validated in both simulation and real-world strawberry-harvesting scenarios, where ILMSA demonstrated a high success rate and practical feasibility. Furthermore, the continuous harvesting control system was validated, confirming its capability to integrate perception, path planning, and harvesting sequence generation. Overall, ILMSA provides an efficient and effective solution for real-time path planning in agricultural robotics, offering great potential for advancing agricultural automation in future research.

Future efforts will focus on designing an end-effector capable of continuously harvesting and storing multiple fruits, as well as deploying path planning algorithms for dual-arm harvesting robots to maximize the practical application of continuous planning algorithms. Additionally, further optimization of ILMSA's performance will be pursued. This includes developing a more advanced collision detection model that better accommodates the irregular shapes of strawberries, as well as refining the path evaluation process to strike an optimal balance between path length, safety, and smoothness based on specific application requirements. Furthermore, integrating cutting-edge reinforcement learning techniques and employing adaptive parameter tuning approaches could enhance the algorithm's versatility and adaptability to a wider range of scenarios.

\bibliographystyle{unsrt}
\bibliography{refer}

\end{document}